\documentclass[10pt,twocolumn,letterpaper]{article}

\usepackage{cvpr}
\usepackage{times}
\usepackage{epsfig}
\usepackage{graphicx}
\usepackage{amsmath}
\usepackage{amssymb}

\usepackage{amsmath,amsfonts,bm}

\def\eqref#1{equation~\ref{#1}}

\def\1{\bm{1}}

\DeclareMathAlphabet{\mathsfit}{\encodingdefault}{\sfdefault}{m}{sl}
\SetMathAlphabet{\mathsfit}{bold}{\encodingdefault}{\sfdefault}{bx}{n}

\usepackage{color,xcolor}
\usepackage{array}
\usepackage{booktabs}
\usepackage{colortbl}
\usepackage{float,wrapfig}
\usepackage{hhline}
\usepackage{multirow}
\usepackage{subcaption} %
\usepackage[font={small}]{caption}

\usepackage{amsmath,amsfonts,amsthm,amssymb}
\usepackage{bm}
\usepackage{nicefrac}
\usepackage{microtype}
\usepackage{mathtools}

\usepackage{changepage}
\usepackage{extramarks}
\usepackage{fancyhdr}
\usepackage{lastpage}
\usepackage{setspace}
\usepackage{soul}
\usepackage{xspace}

\usepackage{balance}
\usepackage{pdfpages}
\usepackage{booktabs} %
\newcolumntype{L}[1]{>{\raggedright\let\newline\\\arraybackslash\hspace{0pt}}m{#1}}
\newcolumntype{C}[1]{>{\centering\let\newline\\\arraybackslash\hspace{0pt}}m{#1}}
\newcolumntype{R}[1]{>{\raggedleft\let\newline\\\arraybackslash\hspace{0pt}}m{#1}}

\newcommand{\sect}[1]{Section~\ref{#1}}

\newcommand{\eqn}[1]{Equation~\ref{#1}}
\newcommand{\fig}[1]{Fig.~\ref{#1}}
\newcommand{\tbl}[1]{Table~\ref{#1}}

\newcommand{\ignore}[1]{}

\makeatletter
\DeclareRobustCommand\onedot{\futurelet\@let@token\@onedot}
\def\@onedot{\ifx\@let@token.\else.\null\fi\xspace}

\def\eg{e.g\onedot} 
\def\ie{i.e\onedot} 
 
\def\etc{etc\onedot}

\def\wrt{w.r.t\onedot}

\def\etal{et al\onedot}
\def\aka{a.k.a\onedot}
\makeatother

\definecolor{MyDarkBlue}{rgb}{0,0.08,1}
\definecolor{MyDarkGreen}{rgb}{0.02,0.6,0.02}
\definecolor{MyDarkRed}{rgb}{0.8,0.02,0.02}
\definecolor{MyDarkOrange}{rgb}{0.40,0.2,0.02}
\definecolor{MyPurple}{RGB}{111,0,255}
\definecolor{MyRed}{rgb}{1.0,0.0,0.0}
\definecolor{MyGold}{rgb}{0.75,0.6,0.12}
\definecolor{MyDarkgray}{rgb}{0.66, 0.66, 0.66}

\newcommand{\problemfull}{perspective plane program induction\xspace}
\newcommand{\Problemfull}{Perspective Plane Program Induction\xspace}
\newcommand{\model}{P3I\xspace}
\newcommand{\program}{perspective plane program\xspace}

\newcommand{\programs}{perspective plane programs\xspace}
\newcommand{\Programs}{Perspective Plane Programs\xspace}
\newcommand{\dataset}{NRPP\xspace}
\newcommand{\datasetfull}{Nearly-Regular Patterns with Perspective\xspace}

\newcommand{\mysubsection}[1]{\vspace{-4pt}\subsection{#1}\vspace{-4pt}}
\newcommand{\myparagraph}[1]{\vspace{-14pt}\paragraph{#1}}

\newcommand{\thetitle}{Perspective Plane Program Induction from a Single Image}

\usepackage[pagebackref=true,breaklinks=true,letterpaper=true,colorlinks,bookmarks=false]{hyperref}

\cvprfinalcopy %

\def\cvprPaperID{5738} %

\ifcvprfinal\fi
\begin{document}

\title{\thetitle}

\author{
Yikai Li$^{1,2*}$\qquad Jiayuan Mao$^{1*}$\qquad Xiuming Zhang$^{1}$\\ 
William T. Freeman$^{1,3}$\qquad Joshua B. Tenenbaum$^{1}$\qquad Jiajun Wu$^{4}$\\
\vspace*{3pt}
$^1$MIT CSAIL\qquad $^2$Shanghai Jiao Tong University\qquad $^3$Google Research\qquad $^4$Stanford University\\
\vspace*{-10pt}
}

\thispagestyle{empty}

\twocolumn[{%
\renewcommand\twocolumn[1][]{#1}%
\vspace{-2em}
\maketitle

\vspace{-1.5em}
    \centering
    \includegraphics[width=\textwidth]{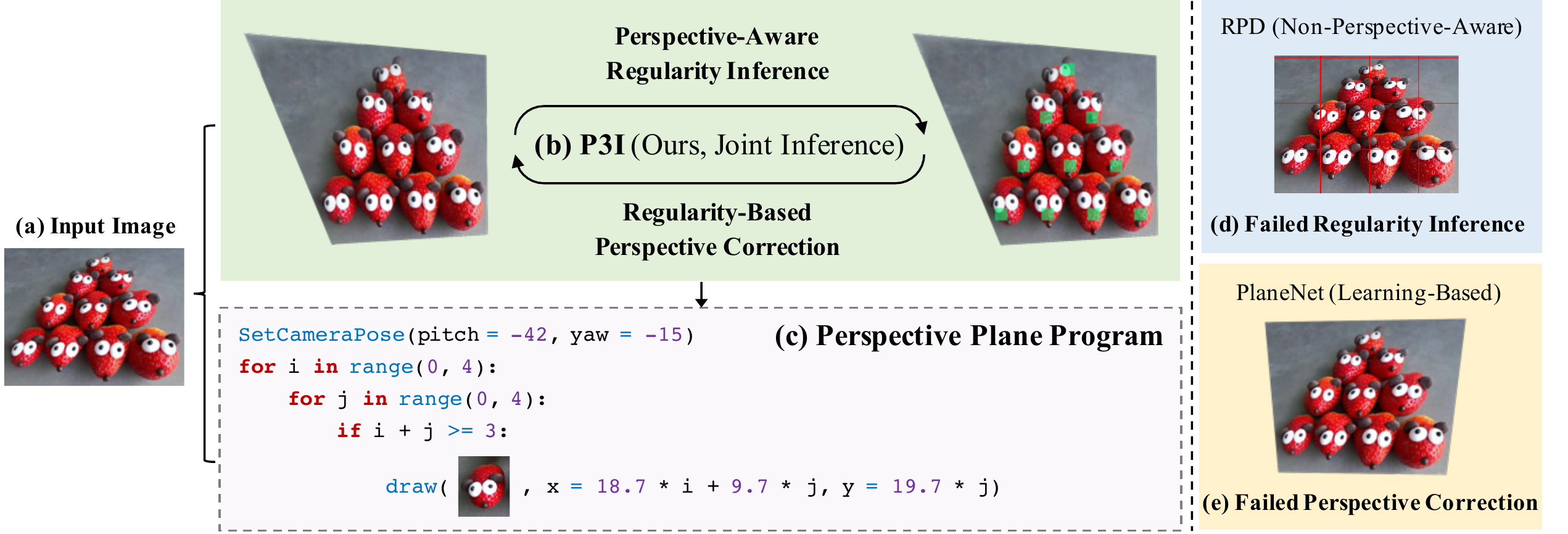}
    \vspace{-15pt}
    \captionof{figure}{Perspective effects and scene structure regularity are ubiquitous in natural images (a). To detect such regularity, one may directly apply regularity structure detection (RPD)~\cite{lettry2017repeated} to natural images, but this often fails due to the existence of perspective effects (d). 
    Attempting to remedy this, one may perform perspective correction as an independent preprocessing step, but perspective correction often relies on line and/or vanishing point cues, and fails when such cues are missing (e). We observe that these two tasks are interconnected: image regularity serves as a new perspective correction cue, and regularity detection, in turn, also benefits from perspective correction. %
    Thus, we propose to jointly solve perspective correction and regularity structure detection (b) by simultaneously seeking the program and perspective parameters that best describe the image (c). Project page: \url{http://p3i.csail.mit.edu}
    }
    \label{fig:teaser}
\vspace{1em}
}] 
\begin{abstract}
   We study the inverse graphics problem of inferring a holistic representation for natural images. Given an input image, our goal is to induce a neuro-symbolic, program-like representation that jointly models camera poses, object locations, and global scene structures. Such high-level, holistic scene representations further facilitate low-level image manipulation tasks such as inpainting. We formulate this problem as jointly finding the camera pose and scene structure that best describe the input image.
   The benefits of such joint inference are two-fold: scene regularity serves as a new cue for perspective correction, and in turn, correct perspective correction leads to a simplified scene structure, similar to how the correct shape leads to the most regular texture in shape from texture.
   Our proposed framework, \Problemfull (\model), combines search-based and gradient-based algorithms to efficiently solve the problem. \model outperforms a set of baselines on a collection of Internet images, across tasks including camera pose estimation, global structure inference, and down-stream image manipulation tasks.
\end{abstract} 
\thispagestyle{empty}
\maketitle

\section{Introduction}
\vspace{-0.5em}
\footnotetext{$^*$ indicates equal contribution.}

From a single image in \fig{fig:teaser}, humans can effortlessly induce a holistic scene representation that captures both local textures and global scene structures. We can localize the objects in the scene (the ``strawberry mice'').
We also see the global scene {\it regularities}: the mice collectively form a 2D lattice pattern with a triangular boundary. Meanwhile, we can estimate the camera pose: the image is shot 
at an elevation of roughly 45 degrees.

Building holistic scene representations requires scene understanding from various perspectives and levels of detail: estimating camera poses~\cite{liu2018geometry,bruls2019right,abbas2019geometric}, detecting objects in the scene~\cite{he2017mask,lettry2017repeated}, and inferring the global structure of scenes~\cite{ellis2018learning,mao2019program}. Humans are able to resolve these inference tasks simultaneously. The estimation of global camera pose guides the localization of individual objects and the summarization of the scene structure, such as the lattice pattern in \fig{fig:teaser}(a), in a top-down manner. Meanwhile, the localization of individual objects provides bottom-up cues for the inference of both scene structures and camera poses.

While various algorithms have been developed to tackle each individual task, there is still a lack of studies on the integration of these methods and how they can benefit from each other.
In this paper, we present the framework, \Problemfull (\model), for the joint inference of the camera pose, the localization of individual objects, and a program-like representation that describes lattice or circular regularities of object placement. The inferred holistic scene representation, namely the \program, has a program-like structure with continuous graphics parameters. The key assumption is that the image, possibly captured with perspective effects, is composed of a collection of similar objects that are placed following a {\it regular} pattern.

The integrated inference has three advantages. First, conventional estimations of camera poses (specifically the 3D rotations)
mainly rely on geometric cues, such as straight lines~\cite{tardif2009non,abbas2019geometric} and manually designed texture descriptors \cite{acm_autoRectify_12}, or learning from human annotations ~\cite{liu2018planenet}. Thus, they fail when no straight lines or textual regions can be detected and exhibit poor generalization to unseen complex scenes. %
In this work, \model exploits regular structures on 2D planes to accurately estimate the camera pose. For example, in \fig{fig:teaser}(b), the estimated camera pose can perspective correct the image such that all adjacent mice share roughly the same displacement.
Second, classic object localization algorithms mostly rely on human heuristics~\cite{uijlings2013selective,zitnick2014edge} or require large-scale datasets~\cite{he2017mask}. In this paper, we present a complementary solution based on the similarity among objects in a single image and the global scene {\it regularity}. Such regularities are modeled with the proposed \programs.

Third, although graphics programs, as shown in \fig{fig:teaser}(c), have been found useful for both low-level manipulation and high-level reasoning tasks~\cite{wu2017neural,ellis2018learning,liu2019learning}, the inference is usually not done in an end-to-end manner. These methods work on estimated or known camera parameters and object detection results by off-the-shelf tools, and formulate the inference problem as a pure program synthesis problem in a symbolic space. This restricts the applicability of these algorithms to natural images. By contrast, in this work, \model removes such dependencies by formulating the whole problem as a joint inference task of the camera pose, object locations, and the global scene structure. We show that our model can infer holistic \programs from a single input image without extra tools for any of the tasks.

We collect a dataset of Internet images, namely the \datasetfull dataset (\dataset), for evaluation. The dataset contains non-fronto-parallel images that are composed by a set of objects organized in regular patterns.
\model is evaluated on \dataset in two metrics: accuracy of camera pose estimation and that of graphics programs. Our model outperforms all baselines that tackle these problems separately. Moreover, we show how such holistic representations can be used to perform lower-level image manipulation tasks such as image inpainting and extrapolation. Our approach outperforms both learning-based and non-learning-based baselines designed for such tasks. 
\section{Related Works}
\vspace{1em}
\myparagraph{Camera pose estimation and shape from texture.}
The idea of inferring camera poses (the perspective angles) from regularity draws deep connection to the classic work on shape from texture, dated back to the 80's~\cite{blostein1989shape,aloimonos1988shape,malik1997computing,ohta1981obtaining}. The key assumption here is the uniform density assumption (texels are uniformly distributed). Thus, a perspective view of slanted textured surface will show systematic changes in texture density, area, the aspect ratios. Blostein~\etal\cite{blostein1989shape} and Aloimonos~\cite{aloimonos1988shape} recover the slant and tilt of the camera for images containing a single plane, while Malik and Rosenholtz~\cite{malik1997computing} consider curved surfaces. Aiger~\etal\cite{acm_autoRectify_12} finds homography transformations by running statistical analysis on the detected regions of textures. Furthermore, Ohta~\etal\cite{ohta1981obtaining} combines perspective from texture and the estimation of vanishing points. Recently, there have been attempts that leverage deep learning for 3D line, vanishing point, and plane estimation~\cite{liu2018planenet,abbas2019geometric,zhou2019learning}. %
While these methods focus on camera pose estimation, In this work, we propose to jointly tackle the problem with object localization and scene structure prediction via programs.
\begin{figure*}[t]
    \centering
\vspace{-5pt}
    \includegraphics[width=\textwidth]{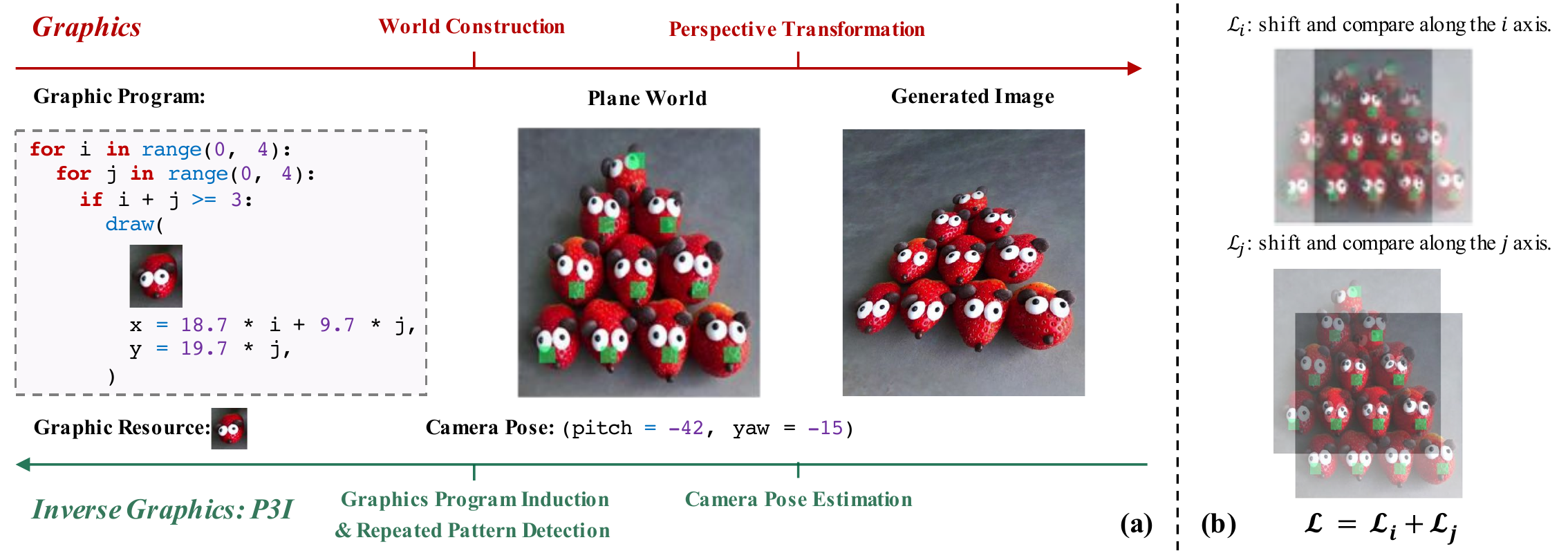}
    \vspace{-2em}
    \caption{\textbf{(a)} Our model \model solves an {\it inverse graphics} problem. Given an input image, \model jointly infers the camera pose, object locations, and the global scene regularity, which is an inversion of a simplified graphics pipeline. \textbf{(b)} We compute the fitness of a program on the image based on a shift-and-compare routine, which we illustrates on a lattice pattern case.
    }
    \vspace{-1.5em}
    \label{fig:model}
\end{figure*} 
\myparagraph{Program induction and inverse graphics.}
Procedural modeling is well-established topic in computer graphics, mostly for indoor scenes~\cite{wang2011symmetry,li2019grains,niu2018im2struct} and 3D shapes~\cite{li2017grass,tian2019learning}. Recently, researchers propose to augment such algorithm with deep recognition networks. Representative works include graphics program induction for hand-drawn images~\cite{ellis2018learning}, 3D scenes~\cite{liu2019learning}, primitive sets~\cite{sharma2018csgnet}, and markup code~\cite{deng2017image,beltramelli2018pix2code}. However, they only work on synthetic images in a constrained domain, while here we study natural images.
SPIRAL~\cite{ganin2018synthesizing}, and its follow-up SPIRAL++~\cite{mellor2019unsupervised}, both used reinforcement learning to discover 'doodles' that are later used to compose the image. Their models are no longer restricted to constrained domains, but are also not as transparent and interpretable as symbolic program-like representations, which limits their applications in tasks that involve explicit reasoning, such as image extrapolation. %

Most relevant to our papers are the work from Young~\etal\cite{young2019learning} and from Mao~\etal\cite{mao2019program}, where they both used formal representations within deep generative networks to represent natural images, and later applied the representation for image editing. Unlike Young~\etal\cite{young2019learning}, which requires learning semantics on a pre-defined dataset of semantically similar images, our \model learns from a single image, following the spirit of internal learning~\cite{shocher2018zero}. Unlike Mao~\etal\cite{mao2019program}, which assumes a top-down view and fails on images with perspective distortions, \model simultaneously infers the camera pose, object locations, and scene structures.

\myparagraph{Image manipulation.}
Image manipulation is most commonly studied in the context of image inpainting. Inpainting algorithms can be based on pixels, patches, or global image representations. Pixel-based methods~\cite{ashikhmin2001synthesizing,ballester2001filling} and patch-based methods~\cite{efros2001image,barnes2009patchmatch} perform well when the missing regions are small and local, but cannot deal with cases that require high-level information beyond background textures.
Darabi~\etal\cite{Darabi2012ImageMC} extended patch-based methods by allowing additional geometric and photometric transformations on patches but ignored global consistency among patches. Huang~\etal\cite{Huang2014ImageCU} also used perspective correction to help patch-based inpainting, but their algorithm relies on the vanishing point detection by other methods. By contrast, \model estimates the camera parameters based on the global regularity of images.

The advances of deep nets has led to many impressive inpainting algorithms that integrate information beyond local pixels or patches~\cite{iizuka2017globally,yang2017high,yu2018generative,liu2018image,yu2019free,zhou2018non,yan2018shift}. Most relevant to our work, Xiong~\etal~\cite{xiong2019foreground} and Nazeri~\etal~\cite{nazeri2019edgeconnect} proposed to explicitly model contours to help the inpainting system preserve global object structures. Compared with them, \model manipulates the image based on its latent \program. Thus, we can preserve the global scene {\it regularity} during manipulation and requires no extra training images. %
\section{\Problemfull}

The proposed framework, \Problemfull (\model), takes a raw image as input and infers a \program that best describes the image. In \sect{sec:dsl}, we first present the domain-specific language of the program that we use to describe the scene and the camera, by walking through a graphics pipeline that generates a natural image. In \sect{sec:model}, we present our algorithm for the inversion of such \programs and mathematically formulate it as a joint inference problem. Finally, in \sect{sec:inference}, we present a hybrid inference algorithm to perform the inference efficiently. Implementation details are supplied in \sect{sec:implementation}.

\begin{figure}[t!]
    \centering
    \includegraphics[width=\linewidth]{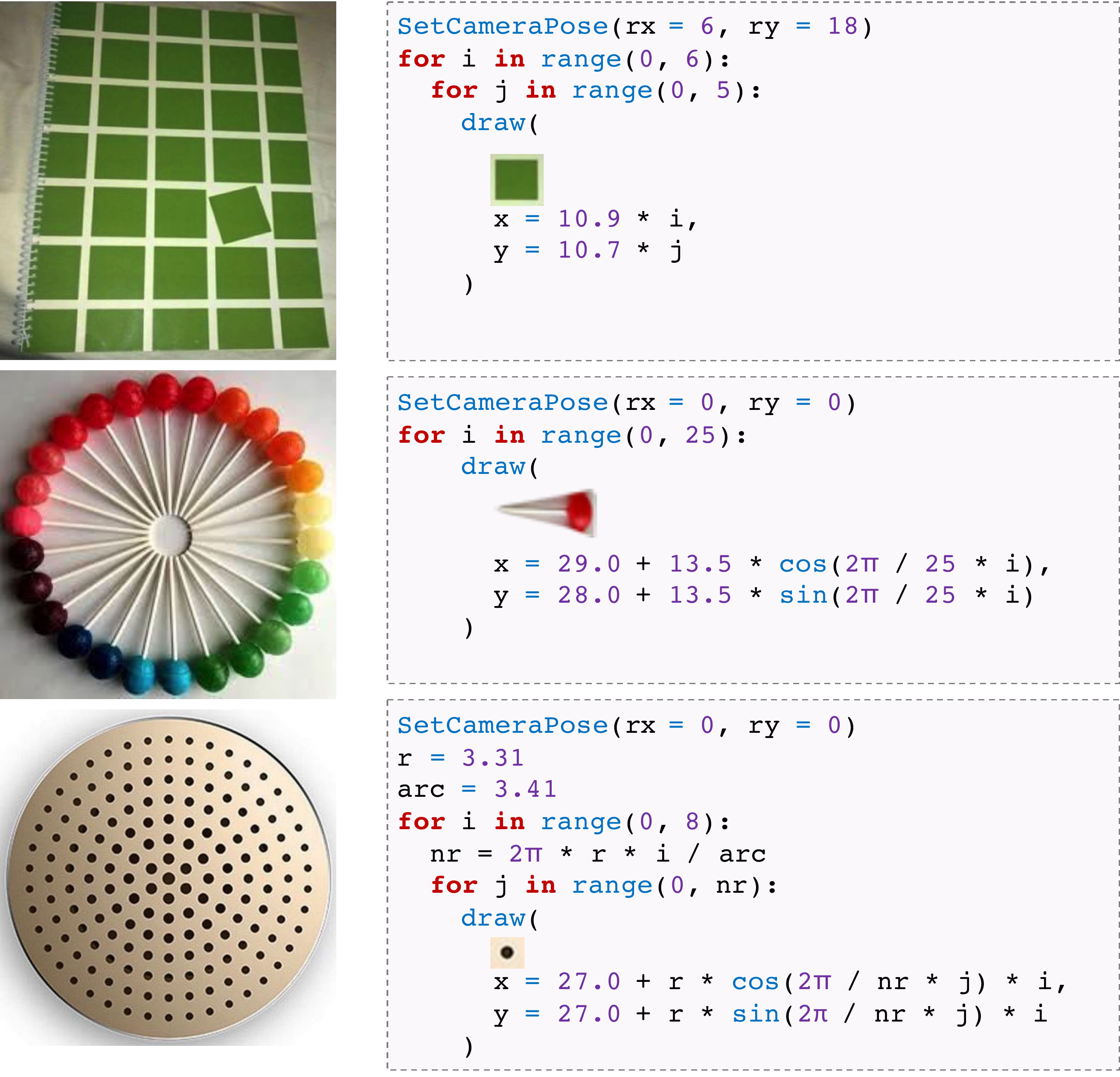}
    \caption{Example programs inferred by \model. Our model can perform joint inference of camera pose, object localizations, and global scene structures of images having different regularity patterns: (a) lattice, (b) circular, and (c) a hybrid structure composed by a circular structure and a linearly repeated one.}
    \vspace{-1em}
    \label{fig:programs}
\end{figure} 
\mysubsection{\Programs}
\label{sec:dsl}
We introduce our \programs by walking through the graphics pipeline that generates a natural image. Suppose that the scene is composed of a collection of visually similar objects (or generally, patterns) placed regularly on a 2D plane. Thus, the generative process of the image can be divided into three parts: first, modeling of individual objects; second, global scene regularities such as the lattice patterns in \fig{fig:programs} (a) or circular patterns in \fig{fig:programs} (b) and (c), represented using graphics programs; and third, camera extrinsic and intrinsic parameters, which defines the projection of the 3D scene onto a 2D image plane.

Illustrated in \fig{fig:programs}, a \program consists of the primitive {\tt Draw} command which places objects at specified positions. Such {\tt Draw} commands appear in (possibly nested) {\tt For}-loop and {\tt Rotate}-loop statements, which characterizes the global scene regularity. Finally, the program specifies camera parameters with the command {\tt SetCameraPose}. We restrict the nested loops to be at most two-level because 1) it is powerful enough to capture most 2D layouts, and 2) in perspective geometry, a two-dimensional pattern is sufficient for inferring the vanishing line of a plane (and thus the plane orientation). However, we can expand the DSL to include new patterns; the inference algorithm we present is also not tied to any specific DSL and generalizes to new patterns.

\myparagraph{Repeated patterns.} The most basic command in a \program is {\tt Draw}. Given 2D coordinates $(x, y)$, a single call to the {\tt Draw} command places an object or, generally, a pattern on the 2D plane, centering at $(x, y)$. The {\tt Draw} commands are enclosed in (nested) loops that define lattice or circular structures. \fig{fig:programs} illustrates the latent \programs for a set of images.

\myparagraph{Perspective transformations.} The next step of the graphics pipeline is to project the 3D space onto a 2D image plane. Since we consider only a single 2D plane in the 3D world, the resulting transformation can be modeled as a perspective transformation (which gives the name, \programs). For simplicity, we only model the 3D rotational transformations given by the camera pose and make a set of assumptions on the other intrinsic and extrinsic parameters of the camera. Details could be found in \sect{sec:implementation}.

\mysubsection{Inversion of the Graphics Pipeline}
\label{sec:model}
The goal of \model is to solve an inverse graphics problem: given the generated image of the scene, we want to estimate the camera pose, infer the regular pattern, and localize the individual objects or patterns. We view this problem as finding a program $P$ that best fit the input image $I$. In this section, we demonstrate how our fitness function is computed. The backward direction of \fig{fig:model} gives an illustration.

Taking the RGB image as the input, we first extract its visual feature using an ImageNet-pretrained AlexNet~\cite{Krizhevsky2012Imagenet}. Working on the feature space makes the inference procedure more robust to local noises such as luminance and reflectance variations, compared with working with RGB pixels directly. We denote $\mathcal{F}_{\mathrm{AlexNet}}$ as the feature extractor and $F = \mathcal{F}_{\mathrm{AlexNet}}(I)$ as the extracted visual features.

The second step is to invert the 2D projection. Assuming a pin-hole camera model, this is done by performing an inverse perspective transformation on the feature $F$. Specifically, we transform the extracted feature map as a fronto-parallel feature based on the XYZ rotations $\mathit{rx}, \mathit{ry}, \mathit{rz}$:
{\small
\small\vspace{-5pt}
\begin{equation}
F^{\mathit{fp}} = \mathrm{WarpPerspective}_{-rx, -ry, -rz}(F).    
\end{equation}
}
Note that, ideally, the transformation should be done on the input image. However, in practice, we swap the order of perspective transformation and AlexNet feature extraction. We find that transforming the feature map provides a good approximation of extracting features on the transformed image, \ie
{\small
\small\vspace{-5pt}
\begin{eqnarray*}
& & \mathrm{WarpPerspective}_{-rx, -ry, -rz}(\mathcal{F}_{\mathrm{AlexNet}}(I))\\
& \approx & \mathcal{F}_{\mathrm{AlexNet}} \big( \mathrm{WarpPerspective}_{-rx, -ry, -rz}(I) \big).
\end{eqnarray*}}
Moreover, performing feature map transformation is more computationally efficient: we do not need to run the AlexNet multiple times for different camera parameters.

The next step is to reconstruct the scene structure and localize individual objects. This is formulated as synthesizing a program that describes the transformed canvas plane. Each candidate program in the DSL space produces a set of 2D coordinates that can be interpreted as centers of objects. We compute the loss of each program based on the similarity of objects that are located by the program.

Mathematically, we denote $C$ as a set of 2D coordinates generated by a program graphics program $P$, defined on the (transformed) canvas plane. Since a \program contains at most a two-level nested loops, we denote the loop variables as $i$ and $j$ and view each coordinates $(x, y)$ in $C$ as a function of the loop variables. That is, $(x, y) = (x(i, j), y(i, j))$. The coordinate functions can be chosen to fit either lattice or circular patterns. We define the loss function as:

{
\footnotesize \vspace{-5pt}
\begin{align}
\mathcal L = & \sum_{i, j} \left\|F^{\mathit{fp}}[x(i, j), y(i, j)] - F^{\mathit{fp}}[x(i+1, j), y(i+1, j)] \right\|_2^2 \nonumber\\
+ & \sum_{i, j} \left\|F^{\mathit{fp}}[x(i, j), y(i, j)] - F^{\mathit{fp}}[x(i, j+1), y(i, j+1)] \right\|_2^2\label{eq:loss},
\end{align}
\normalsize
}
where $\|\cdot\|_2$ is the $\mathcal{L}_2$-norm, which computes the difference between two feature vectors at two spatial positions. In the lattice case, illustrated in \fig{fig:model}(b), one can interpret this fitness function as the following operations: we shift the feature map by a displacement; we then compute the feature similarity between the shifted and the original feature map.

In contrast to previous work by Lettry~\etal\cite{lettry2017repeated}, which detects repeated patterns based on a lattice global structure assumption, our program-based formulation allows a more flexible and compositional way to define global scene structures and perform inference. As an example, in \fig{fig:programs}(c), our model can detect repeated patterns in a hybrid structure composed by a circular structure and a linearly repeated one.

\mysubsection{Grid Search and Gradient-Based Optimization}
\label{sec:inference}

We present a hybrid inference algorithm to solve the problem of finding a graphics program $P$ that best fits the input image $I$. The output of the algorithm includes both the layout patterns (lattice, circular, \etc) and the parameters. This requires solving an optimization problem of choosing a discrete structure (\eg, lattice or circular) and a collection of continuous parameters (rotation angles, object locations, \etc). Previously, graphics program inference is tackled mostly by program synthesis via search in the symbolic program space~\cite{ellis2018learning,mao2019program}. This search process is slow, because the symbolic space is huge, growing exponentially with respect to the number of parameters. It is often required to quantize parameters (\eg, to integers) and use heuristics to accelerate the search process. Unlike these approaches, \model tackles such inference with a hybrid version of search-based and gradient-based optimization.

The key insight here is that both WarpPerspective transformations and feature map indexing are differentiable \wrt the parameters (the rotational angles and the continuous 2D coordinates), since they are both implemented using bilinear interpolation on the feature maps.
Thus, the loss function (\eqn{eq:loss}) is differentiable with respect to all parameters in \programs, including camera poses and constants in coordinate expressions, making gradient-based optimization applicable to our inference task.

\begin{figure}[t]
    \centering
    \includegraphics[width=\linewidth]{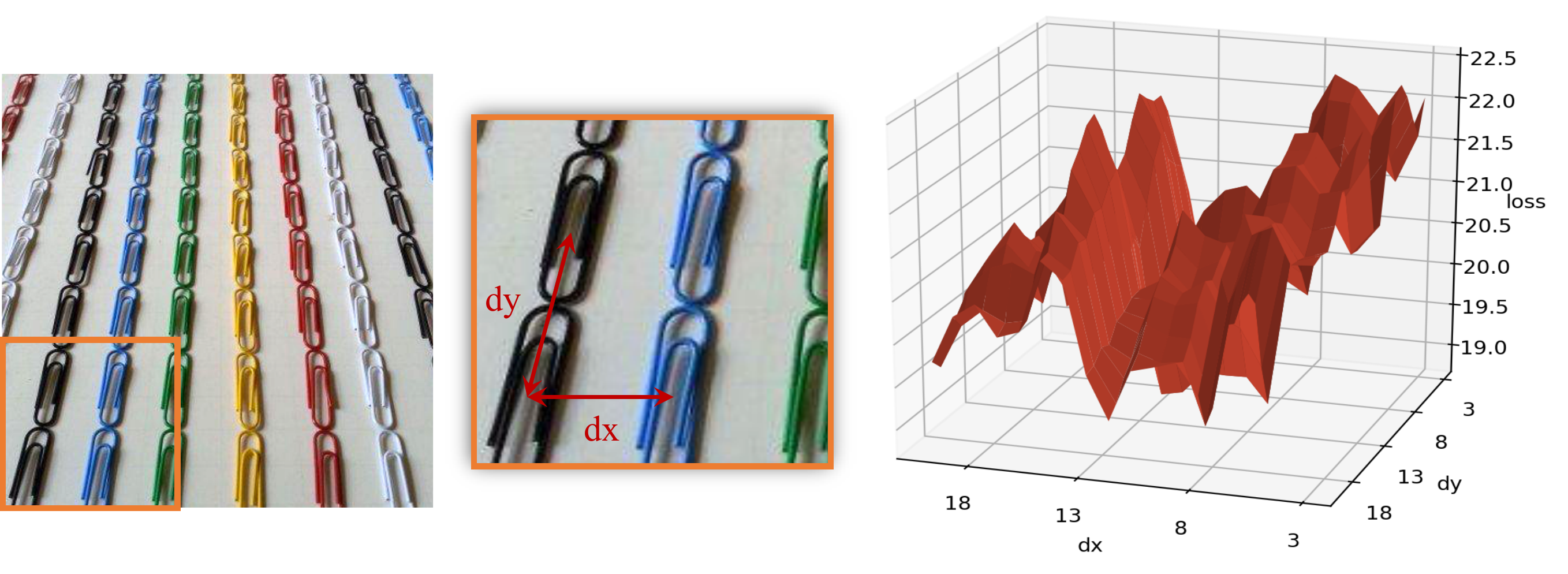}
    \vspace{-2em}
    \caption{Loss Surface of the displacement parameter in a \program. On the left we show the input image. The regularity in the image directly leads to many cliffs on the loss surface ($dx = 10, 12.5, 15, \cdots$) and many local peaks.
    }
    \vspace{-1.5em}
    \label{fig:loss_surface}
\end{figure}
 
However, directly applying gradient descent on $\mathcal L$ remains problematic: the discrete nature of object placements makes the loss function (\eqn{eq:loss}) non-convex. As shown in \fig{fig:loss_surface}, the regularity in the image leads to many cliffs and peaks (local optima). Therefore, direct application of gradient descent will get stuck at a local optimum easily.

Thus, we propose a hybrid inference algorithm to exploit the robustness of search-based inference and the efficiency of gradient-based inference. Specifically, we perform discrete search on the choice of regularity structure. Three structures are considered in this paper, as illustrated in \fig{fig:programs}: (a) lattice, (b) circular, and (c) a hybrid one. For continuous parameters, we perform grid search on a coarse scale and apply fine-grained gradient descent only locally. This is simply implemented by perform a grid search of initial parameters and performing gradient descent on each individual combination.

\mysubsection{Implementation Details}
\label{sec:implementation}
During inference, in the grid search, the grid size for the coarse search of continuous values is 2. For lattice patterns, we do not perform search on the boundary conditions for the loop variables. Instead, the boundaries are generated based on the size of the image. In other words, we assume that the regular pattern covers the whole image plane.

Throughout the paper we consider a simplified camera model with only two rotational degree of freedom (the X-tilt and the Y-tilt). Thus, we assume that the optic axis is aligned with the image center and there is no Z-axis rotation. This is because the Z-axis rotation has been captured by the object coordinates. For example, with lattice patterns, objects can be placed along axes that are not in parallel to the X and Y axes.  We do not assume a known focal length $f$ and aspect ratio $\alpha$. They cannot be recovered unequivocally from a single 2D plane, and different $f$ and $\alpha$ yield to the same perspective correction and image editing results. The results shown in the paper, obtained with $f=35$ and $\alpha=1$, will remain the same with other $f$ and $\alpha$. Meanwhile, we ignore lens distortions, such as radial distortion. Our method can be integrated with camera calibration algorithms to correct them based on detected repeated patterns~\cite{Devernay1995Automatic}
\vspace{-4pt}
\section{Experiments}
\vspace{-4pt}

We test our model on a newly collected dataset, \datasetfull (\dataset), and evaluate its accuracy for camera pose estimation (\sect{sec:expr:perspective}) and repeated pattern detection (\sect{sec:expr:rpd}). We further demonstrate the model can be used to guide low-level image manipulation (\sect{sec:expr:image}).

\mysubsection{Dataset}
\label{sec:expr:dataset}

\begin{figure*}[!thbp]
    \centering
    \vspace{-2em}
    \includegraphics[width=\textwidth]{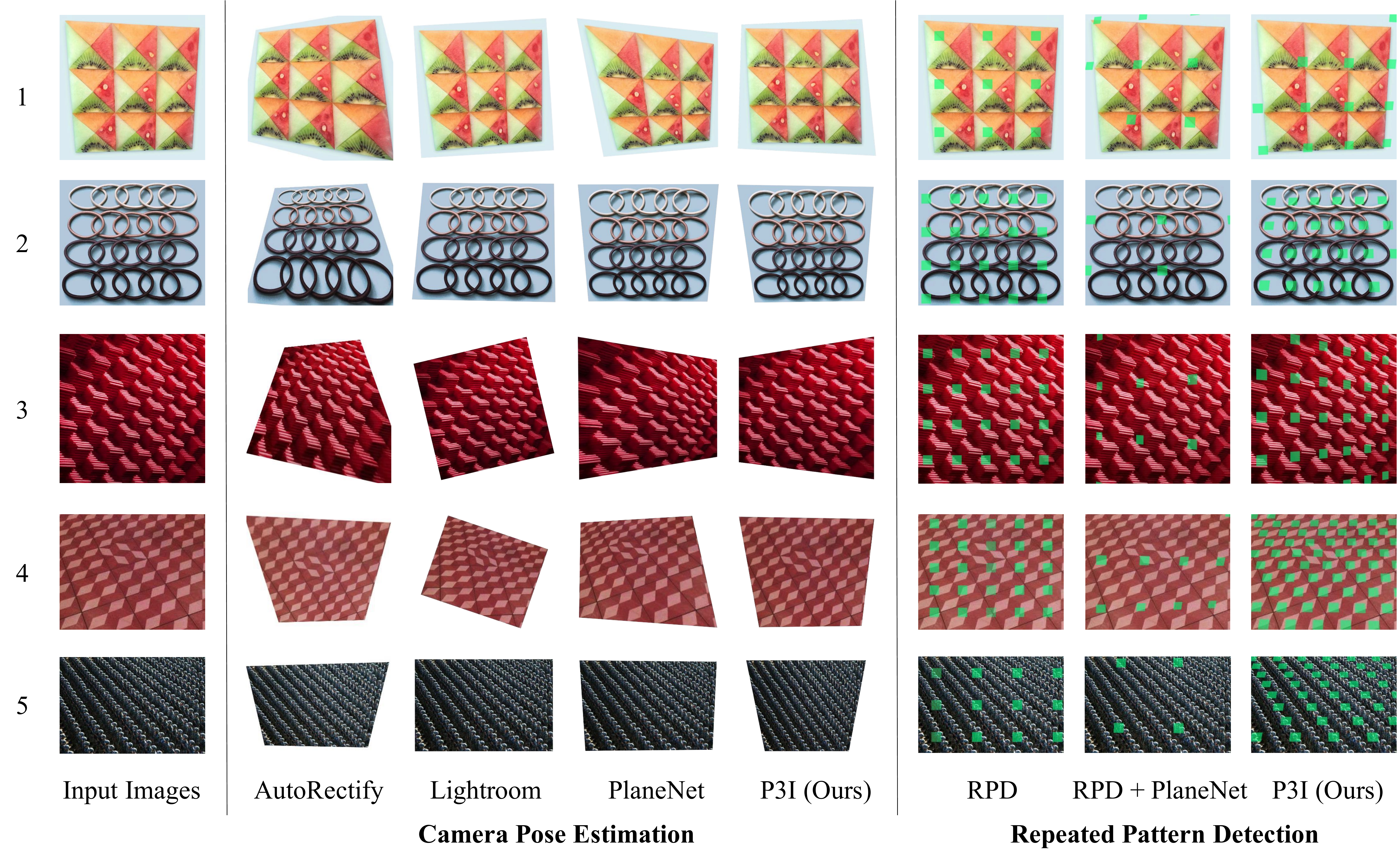}
    \vspace{-2.5em}
    \caption{On the left, we show that \model estimates the camera pose based on the global regularity of images. It outperforms AutoRectify, Lightroom and PlaneNet. Results are visualized by performing an perspective correction based on the estimated parameters. On the right, we show that \model can perform perspective-aware repeated pattern detection, while both RPD~\cite{lettry2017repeated} and RPD+PlaneNet fail.}
    \vspace{-1em}
    \label{fig:qresult_rpd}
\end{figure*}

We collected a dataset of 64 Internet images that each contain a set of objects organized in regular patterns (lattice and circular). Unlike a similar dataset, Nearly Regular Patterns~\cite{lettry2017repeated}, all images in our \dataset are not fronto-parallel; \fig{fig:qresult_rpd} gives some examples. We augment the dataset with human annotations of the camera pose and object locations, in the form of 2D coordinates of object centers. This supports a quantitative evaluation for camera pose estimation and repeated object detection.

\mysubsection{Camera Pose Estimation}
\label{sec:expr:perspective}

We evaluate the performance of \model on camera pose estimation from single images, compared against both learning-based and non-learning-based baselines. %

\myparagraph{Baselines.}
We compare \model with three baselines. The first is AutoRectify~\cite{acm_autoRectify_12}, a texture-based baseline for camera pose estimation. AutoRectify statistically find homography transformation from intersects of detected ellipse regions. We decompose the output transformation matrix to get camera pose as the prediction of AutoRectify. The second is PlaneNet~\cite{liu2018planenet}, a learning-based baseline for camera pose estimation. PlaneNet is a convolutional neural network-based algorithm trained to detect 2D planes and their normals in 3D scenes from RGB images. Since all images in our dataset contain only one plane, we select the largest plane detected by PlaneNet and use its normal vector to compute the camera pose as the prediction of PlaneNet.  Across all experiments, we use the PlaneNet model pretrained on ScanNet~\cite{dai2017scannet}. We also compare our results qualitatively with the auto-perspective tool provided by Adobe Lightroom.

\myparagraph{Metrics.}
We evaluate the accuracy of the estimated camera poses, \ie, the camera orientation, by calculating their $\mathcal{L}_1$ distance to the human-annotated pose using Rodrigues’ rotation formula. All error metrics are computed in degrees and averaged over all images in the dataset.

\begin{table}[t!]
    \small \centering
    \begin{tabular}{lccc}
        \toprule
        Method & Camera Pose Error\\
        \midrule
        AutoRectify~\cite{acm_autoRectify_12} & 30.54 \\
        PlaneNet~\cite{liu2018planenet} & 23.75 \\
        \model (Ours) & {\bf 4.54}\\
        \bottomrule
    \end{tabular}
    \vspace{-0.5em}
    \caption{Camera pose estimation. \model outperforms texture-based baseline, AutoRectify, and the neural baseline, PlaneNet, by a remarkable margin on the \dataset dataset.
    }
    \vspace{-1.5em}
    \label{tab:result_camera}
\end{table} 
\myparagraph{Results.}
We first present qualitative results in \fig{fig:qresult_rpd}, visualizing the predictions of \model, PlaneNet, and Adobe Lightroom. Our model achieves near-perfect estimations of the camera pose, whereas other baselines lead to incorrect perspective correction, possibly due to the absense of straight line cues. Quantitatively, as shown in \tbl{tab:result_camera}, our model also outperforms AutoRectify and PlaneNet by a significant margin. Since Adobe Lightroom does not provide numerical values of the estimated camera pose, we are unable to make quantitative comparison with it.

These results indicate that our model can successfully use cues from global scene regularities to guide the inference of the camera pose. This differs from traditional visual cues such as vanishing points or straight lines.

\mysubsection{Repeated Pattern Detection}
\label{sec:expr:rpd}

The task of repeated pattern detection is to localize individual objects or patches in an image, assuming that these patches have similar visual appearances.

\myparagraph{Baselines.}
We compare our algorithm with a non-learning-based algorithm, RPD~\cite{lettry2017repeated}, designed for localizing objects that form lattice patterns. We use a subset of \dataset of 56 images (lattice patterns only), each of which contains only lattice patterns. Because the original RPD algorithm is not designed to handle non-fronto-parallel images, we also augment RPD with perspective correction based on the camera pose estimated by PlaneNet.

\myparagraph{Metrics.} The output of both RPD and \model is a list of object centroids, which we compare against the ground-truth annotations of object centroids in the image. Two complementary metrics are used: average distance from all detected centroids to their nearest ground-truth centroids (``Detected to GT'' in \tbl{tab:result_rpd}) and average distance from all ground-truth centroids to their respective closest detected counterparts (``GT to Detected''). Using both metrics penalizes both cases with excessively many detections and those with very few. We also report the sum of two aforementioned asymmetric error metrics, \aka, the Chamfer distance.
\begin{table}[t!]
    \small \centering
    \setlength\tabcolsep{2pt}
    \begin{tabular}{lccc}
        \toprule
        Method & Detected to GT & GT to Detected & Chamfer Dis. \\
        \midrule
        RPD & 0.0971 & 0.0909 & 0.1880 \\
        RPD + PlaneNet & 0.1659 & 0.1013 & 0.2672 \\
        \model (Ours) & {\bf 0.0639} & {\bf 0.0881} & {\bf 0.1520} \\
        \bottomrule
    \end{tabular}
    \vspace{-5pt}
    \caption{Repeated object detection. \model outperforms both RPD and RPD+PlaneNet. The degradation from ``RPD'' to ``RPD + PlaneNet'' is explained by incorrect perspective corrections, which lead to larger errors than not performing the correction at all.}
    \label{tab:result_rpd}
\vspace{-5pt}
\end{table} 
\myparagraph{Results.} Qualitatively visualized in \fig{fig:qresult_rpd}, the original RPD algorithm completely fails when the viewing angle deviates from the fronto-parallel view. The integration of PlaneNet helps correct the perspective effects on certain images, but overall degrades the performance due to large errors when the perspective correction is wrong. As \tbl{tab:result_rpd} shows, our model quantitatively outperforms ``RPD + PlaneNet'' by a large margin, suggesting that our joint inference algorithm is superior to a pipeline directly integrating camera pose estimation and object detection.

\begin{figure}[t]
    \centering
    \vspace{-0.5em}
    \includegraphics[width=0.8\columnwidth]{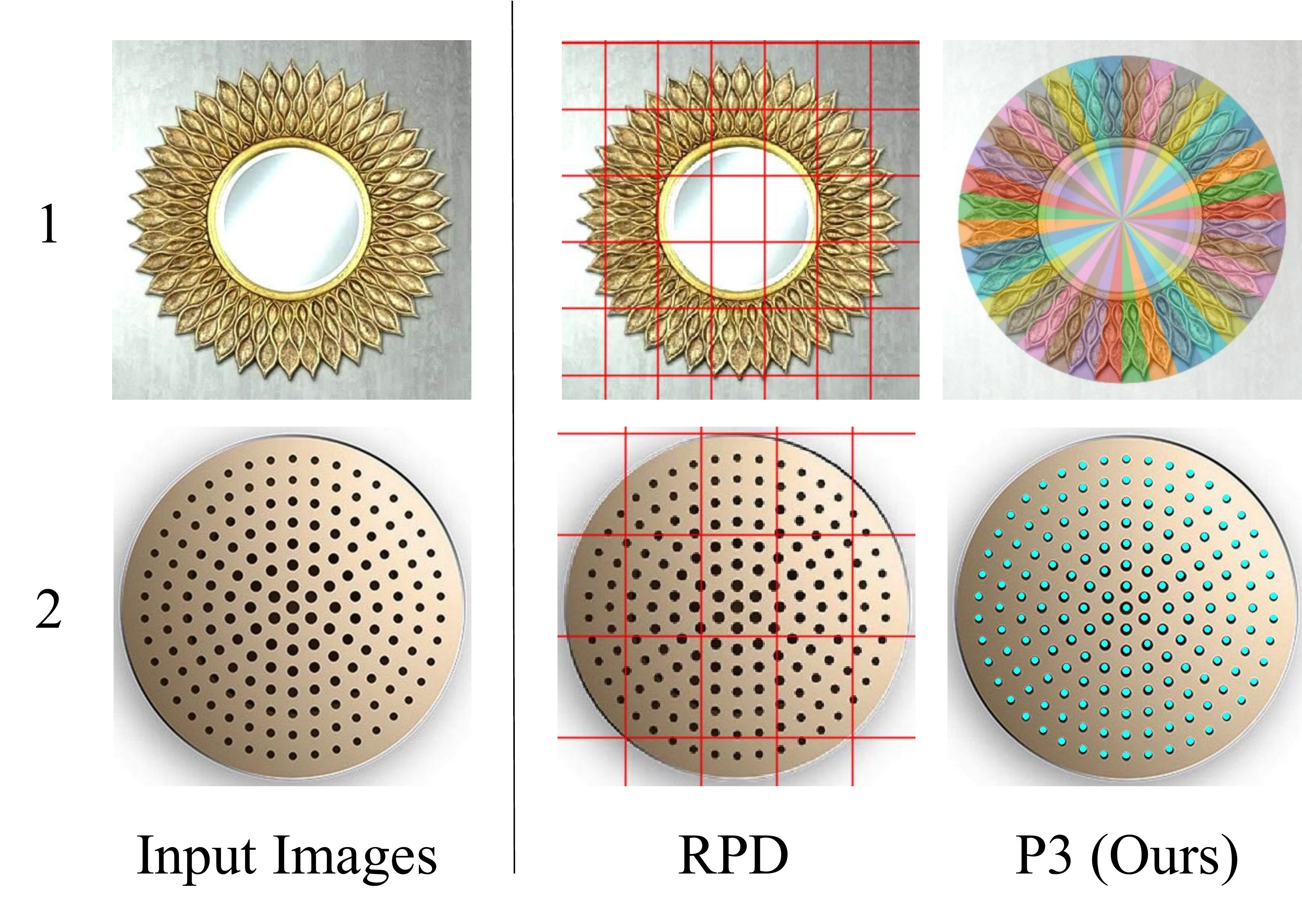}
    \vspace{-1em}
    \caption{Besides lattice patterns, \model is also able to detect (1) circular patterns (rainbow colors are used to visualize the inferred program), and (2) hybrid structures composed by a circular structure and a linearly repeated one.}
    \vspace{-1.5em}
    \label{fig:result_circle}
\end{figure} 
Importantly, \model suggests a general framework for detecting repeated objects organized in {\it any} patterns expressible by a program in the DSL. As an example, we show in~\fig{fig:result_circle} that our model successfully localizes objects with a global circular pattern. Specifically, our model discovers the mirror's radial peripheral and holes on the metal surface.

\mysubsection{Image Manipulation}
\label{sec:expr:image}

The induced perspective plane programs enable perspective-aware image manipulation. The neural painting network (NPN)~\cite{mao2019program} is a general framework for program-guided image manipulation; it performs tasks such as inpainting missing pixels, extrapolating images, and editing image regularities. Consider the representative task of image inpainting. The key idea of NPNs is to train an image-specific neural generative model to inpaint missing pixels based on the specification generated by the high-level program: {\it what} should to be put {\it where}. The original NPNs work only on fronto-parallel images with objects forming lattice patterns.

Thus, we augment the NPN framework to add support for non-fronto-parallel images and non-lattice patterns. Specifically, based on the inferred camera pose, we first transform the input image to a fronto-parallel view. We then train an NPN to manipulate the transformed image. Circular patterns are supported by introducing an extra rotation operation during image manipulation; the details can be found in our supplementary material.

\begin{figure*}[!htbp]
    \centering
    \vspace{-5pt}
    \includegraphics[width=0.9\textwidth]{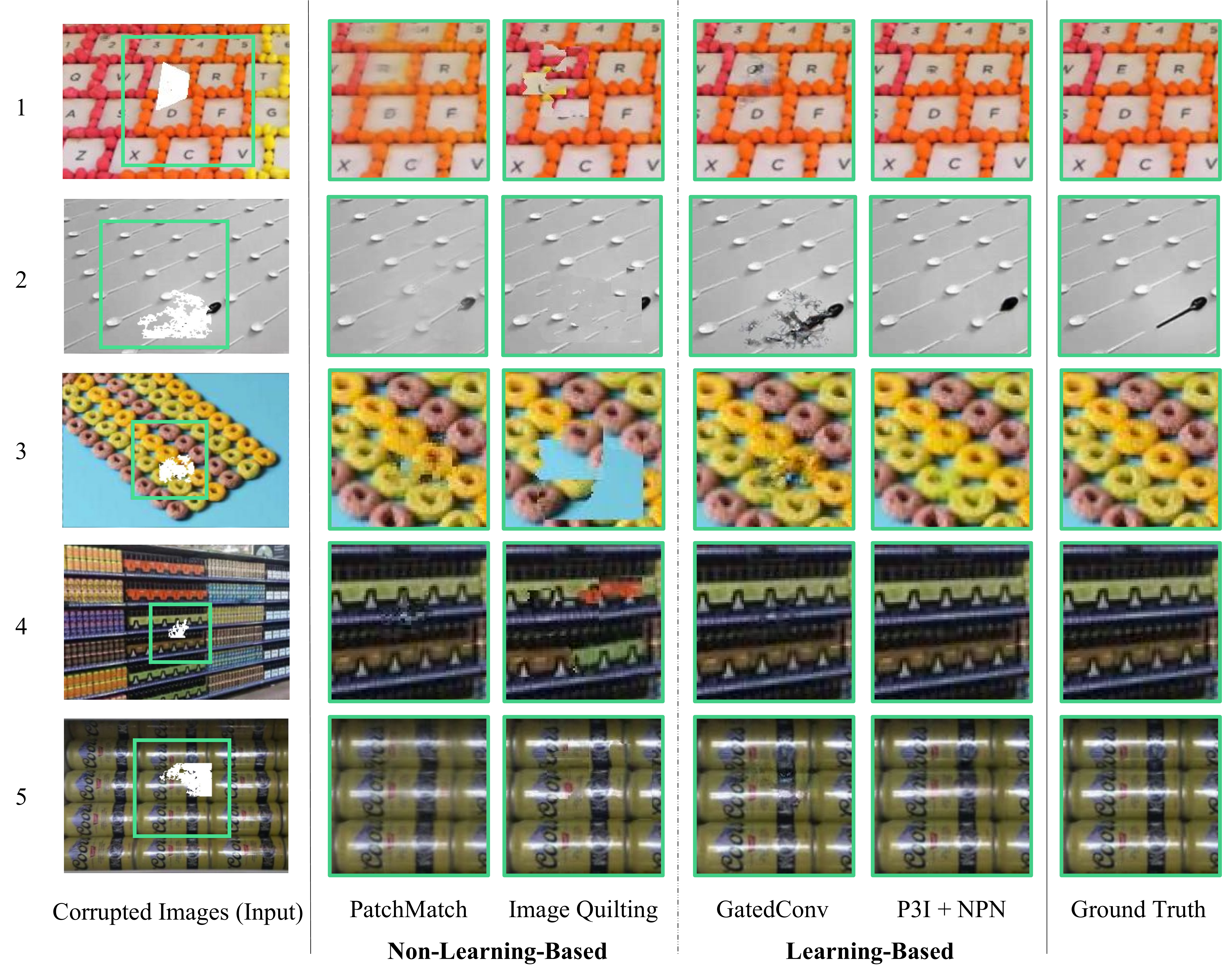}
    \vspace{-5pt}
    \caption{\model-guided NPNs perform inpainting in a perspective-aware fashion. Results generated by \model-guided NPNs are sharp (compared with PatchMatch), consistently connect to the global structure (\eg, the lines in 1 and 4), and respect the global perspective effects.}
    \label{fig:inpaint}
    \vspace{-12pt}
\end{figure*} 
\myparagraph{Baselines.}
We compare our model against both learning-based (GatedConv~\cite{yu2019free}) and non-learning-based algorithms (Image Quilting~\cite{efros2001image} and PatchMatch~\cite{barnes2009patchmatch}). Both non-learning algorithms (Image Quilting and PatchMatch) perform image manipulation based on internal statistics only, without referencing to external image datasets. Similarly, \model-guided NPNs also learn from single images (not external image datasets), and then perform manipulation with learned internal statistics. In contrast, GatedConv is a neural generative model trained on a large collection of images (Places365~\cite{zhou2017places}) for image inpainting.

\myparagraph{Metrics.}
We evaluate the performance of \model-guided NPNs in image inpainting with two metrics: average $\mathcal{L}_1$ distance between the inpainted pixels and ground truth, and Inception Score (IS)~\cite{salimans2016improved} of the inpainted region.

\myparagraph{Results.}
Qualitative results for inpainting are presented in~\fig{fig:inpaint}. Our model can inpaint missing pixels in images at various viewing angles. Both Image Quilting and PatchMatch do not perform well, because they are designed for texture synthesis and assume a stationary texture pattern, and this assumption does not hold when the image is non-fronto-parallel or contains circular patterns. In addition, PatchMatch also modifies pixels near the inpainting region for improved global consistency, resulting in blurriness. More importantly, results by the baselines fail to respect the global perspective pattern (\eg, the lines in 1 and 4). Only \model-guided NPNs are able to inpaint missing objects of proper sizes that respect the overall perspective structure. 

\begin{table}[t!]
    \small \centering
    \begin{tabular}{lcc}
        \toprule
        Method & $\mathcal{L}_1$ & Inception Score \\
        \midrule
        Image Quilting & 35.76  & {\bf 1.16} \\
        PatchMatch & 21.92  & 1.13 \\
        \arrayrulecolor{black!30}\midrule
        GatedConv & 23.20  & 1.12 \\
        \model+NPN (Ours) & {\bf 18.72}  & 1.14 \\
        \arrayrulecolor{black}\bottomrule
    \end{tabular}
    \vspace{-5pt}
    \caption{Image inpainting. \model-Guided NPNs outperform both classic, non-learning baselines and the learning-based baseline in $\mathcal{L}_1$ loss. Image Quilting achieves better Inception Score (IS) than \model-NPNs, because besides patch quality, patch diversity also improves IS~\cite{salimans2016improved}, and Image Quilting tends to produce diverse inpainting across the test images.
    }
    \label{tab:result_inpaint}
\vspace{-15pt}
\end{table} %
\begin{figure}[t]
    \centering
    \vspace{-5pt}
    \includegraphics[width=\columnwidth]{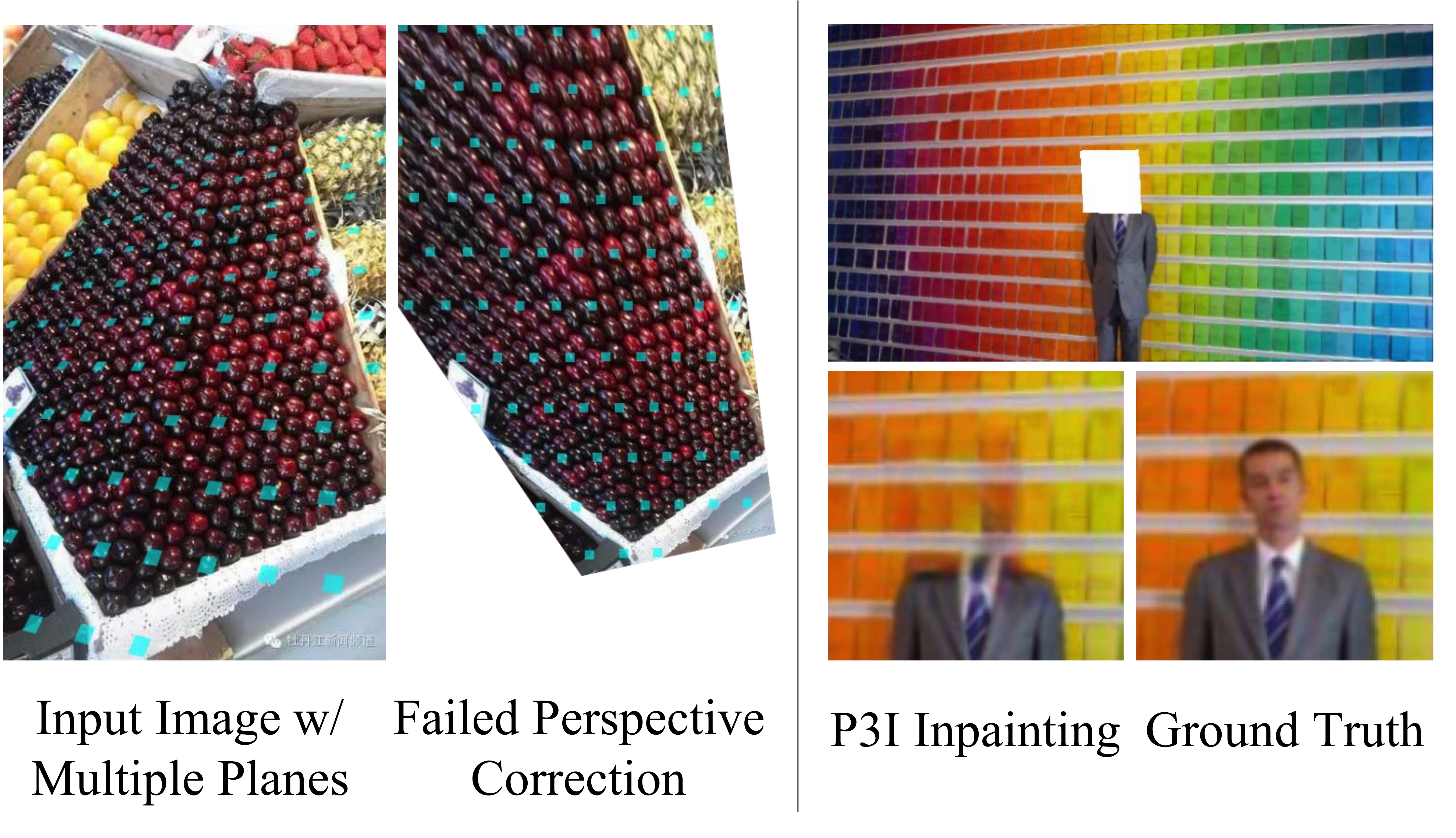}
    \vspace{-20pt}
    \caption{Left: Perspective correction in the presence of multiple planes is a future direction that \model can take. Right: \model inpainting learns low-level texture statistics from single images, so it is unable to inpaint the person's head using high-level semantics.}
    \label{fig:fail}
    \vspace{-10pt}
\end{figure} 
Quantitatively, as shown in \tbl{tab:result_inpaint}, our P3I-guided NPNs outperform all the baselines (both non-learning-based and learning-based) in terms of  $\mathcal{L}_1$ distance between the inpainted pixels and the ground truth. Image Quilting receives a higher Inception Score (IS) than our P3I-NPNs, because high patch diversity, in addition to patch quality, leads to high IS~\cite{salimans2016improved}, and Image Quilting tends to produce diverse inpainting across the test images.

\fig{fig:fail} shows two intriguing failure cases of our model. In the first case (left), the complex scene consists of multiple planes at different orientations, and \model struggles to perspective-correct both planes. In the second case (right), \model only learns low-level texture statistics, therefore failing to inpaint the person's head using high-level semantics.

\vspace{-5pt}
\section{Conclusion}
\vspace{-3pt}

We have presented the \problemfull (\model), a framework for synthesizing graphics programs as a holistic representation for images. A graphics program models camera poses, object locations, and the global scene structure, such as lattice or circular patterns. The algorithm induces graphics programs through a joint inference of the scene structure and camera poses on a single input image, requiring no training or human annotations. A hybrid approach that combines search-based and gradient-based algorithms is proposed to solve the challenging inference task. The induced neuro-symbolic, program-like representations can further facilitate image manipulation tasks, such as image inpainting. The resulting \model-guided neural painting networks (NPNs) are able to inpaint missing pixels in a way that respects the global perspective structure.

\vspace{0.5em}
\noindent {\bf Acknowledgements.}
This work is supported by the Center for Brains, Minds and Machines (NSF STC award CCF-1231216), NSF \#1447476, ONR MURI N00014-16-1-2007, and IBM Research. 
{\small
\bibliographystyle{ieee_fullname}
\bibliography{reference,main}
}
\newpage
\quad
\newpage


\section*{Supplementary material}

# Supplementary Material:
# Perspective Plane Program Induction from a Single Image

The supplementary material is organized as follows. First, in Appendix Section A, we formally defines the domain specific language for perspective plane programs. Next, in Appendix Section B, we discuss the implementation details of neural painting networks (NPNs) used in this paper for image manipulation tasks. Finally, in Appendix Section C, we supplement more experimental results for camera pose estimation, repeated pattern detection, and image manipulation.

## A. Domain Specific Language of Perspective Plane Programs

We formally define the domain specific language (DSL) of perspective plane programs in Table 1. A perspective plane program consists of the primitive Draw command which places objects at specified positions, (possibly nested) For-loop and Rotate-loop statements for characterizing structures, and SetCameraPose commands for set camera poses.

| | | |
|---:|:---:|:---|
| Program | ⟶ | CameraProgram;WorldProgram |
| CameraProgram | ⟶ | SetCameraPose(rx=Real, ry=Real) |
| WorldProgram | ⟶ | For1Stmt \| Rotate2Stmt |
| For1Stmt | ⟶ | For ( $i$ in range(Integer, Integer) ) <br> { For2Stmt \| Rotate2Stmt } |
| For2Stmt | ⟶ | For ( $j$ in range(Integer, Integer) ) <br> { CondDrawStmt } |
| Rotate2Stmt | ⟶ | For ( $j$ in range(ExprI, ExprI) ) <br> { RotateDrawStmt } |
| CondDrawStmt | ⟶ | If (Expr ≥ 0) { CondDrawStmt } |
| CondDrawStmt | ⟶ | DrawStmt |
| DrawStmt | ⟶ | Draw (x=Expr, y=Expr) |
| RotateDrawStmt | ⟶ | Draw (x=RExprX, y=RExprY) |
| Expr | ⟶ | Real * $i$ + Real * $j$ |
| ExprI | ⟶ | Real * $i$ |
| RExprX | ⟶ | Real * Cos(Real * $j$) * $i$ + Real \| Real * Cos(Real * $j$) + Real |
| RExprY | ⟶ | Real * Sin(Real * $j$) * $i$ + Real \| Real * Sin(Real * $j$) + Real |

Table 1: The domain-specific language (DSL) of perspective plane programs. Language tokens For, If, Integer, Real, and arithmetic/logical operators follow the Python convention.

## B. Implementation Details of Neural Painting Networks

The program description of an image provides us with many possible ways of partitioning the image into patches. The most straightforward way may be simply generating the Voronoi diagram with the peaks described by the program. Despite simple, this approach is sufficient to aid downstream image manipulation tasks (performed by neural networks) significantly, as we show in both the main text and this supplementary document.

On a high level, neural networks are known to be good at local pixel manipulations, but less so at understanding more global structure, such as regularity, present in the input image. The program description allows us to aggregate the patches into a "source patch stack" and also align them with the center of the patch to inpaint. By doing so, we reduce a task that originally required understanding the image's global structure into one that requires only inpainting local pixels using a stack of aligned source patches.

Specifically, we adopt the neural painting network (NPN) proposed in [6], a neural network guided by the image program to perform pixel manipulation. For simple illustration, we discuss how NPNs utilize programs in image inpainting only for the

grid programs, but the framework can be easily extended to rotational programs by add a patch rotation step, in addition to the patch translation step.

Here we recap the essense of NPNs for completeness. Interested readers are referred to the original paper [6].

**Patch aggregation.** We first contruct a stack of source patches, from which the NPNs can smartly copy from to inpaint the missing pixels. This is achieved by first generating the Voronoi diagram given the object (loosely defined by the inducted program) centroids, and then breaking down the image into patches according to this Voronoi diagram ("source patches"). We then translate each source patch so that their centers (given by the program) all align with the center of the patch to inpaint. These aligned source patches, together with the corrupted patch with missing pixels, get consumed by a convolutional network and smartly combined into a complete patch with missing pixels filled.

**Architecture.** An NPN is a convolutional network with a U-Net [8] encoder-decoder architecture. Because there are variable numbers of source patches across different input images, it handles an arbitrary number of source patches (as the memory permits). In addition, since the patches form an unordered set, the network is also designed to be invariant to their ordering, producing the same result for different input orders. These two properties are achieved by designs inspired by Aittala et al. [1] and Qi et al. [7], and we refer the reader to the original paper [6] for details. The input of the network is the stack of the corrupted input image plus source patches, and the output of the network is the inpainted image. A detailed printout of the generator's architecture can be found in the supplemental document.

## C. More Experiments and Results

We show three groups of results in this section. First, in Fig. 1 we show the results of our camera pose estimation and repeated pattern detection. Our algorithm P3I outperforms both Lightroom and PlaneNet [5] in camera pose estimation. We find that for images with more salient straight line structures, Lightroom can produce more reasonable results. However, it still fails on most images in our dataset. In contrast, P3I succeeds on most images based on the global regularity cues. As for the repeated pattern detection, our model performs robust perspective-aware detection. Compared with it, the baseline RPD [4] fails when there is a large perspective angle. Adding a PlaneNet-based perspective correction does not show significant improvement (shown as RPD + PlaneNet).

Second, we provide more examples of the induced perspective plane programs from our dataset NRPP. This includes programs for both images with lattice patterns and ones with circular patterns. P3I infers these programs in a unified framework.

Finally, we show more results on the program-guided image inpainting for images with structural regularity and perspective. In general, PatchMatch [2] produces blurry results when operating on non-fronto-parallel images, because this breaks the stationary assumption used by PatchMatch. Both Image Quiting [3] and GatedConv [9] fail to preserve the structural regularities in the image.

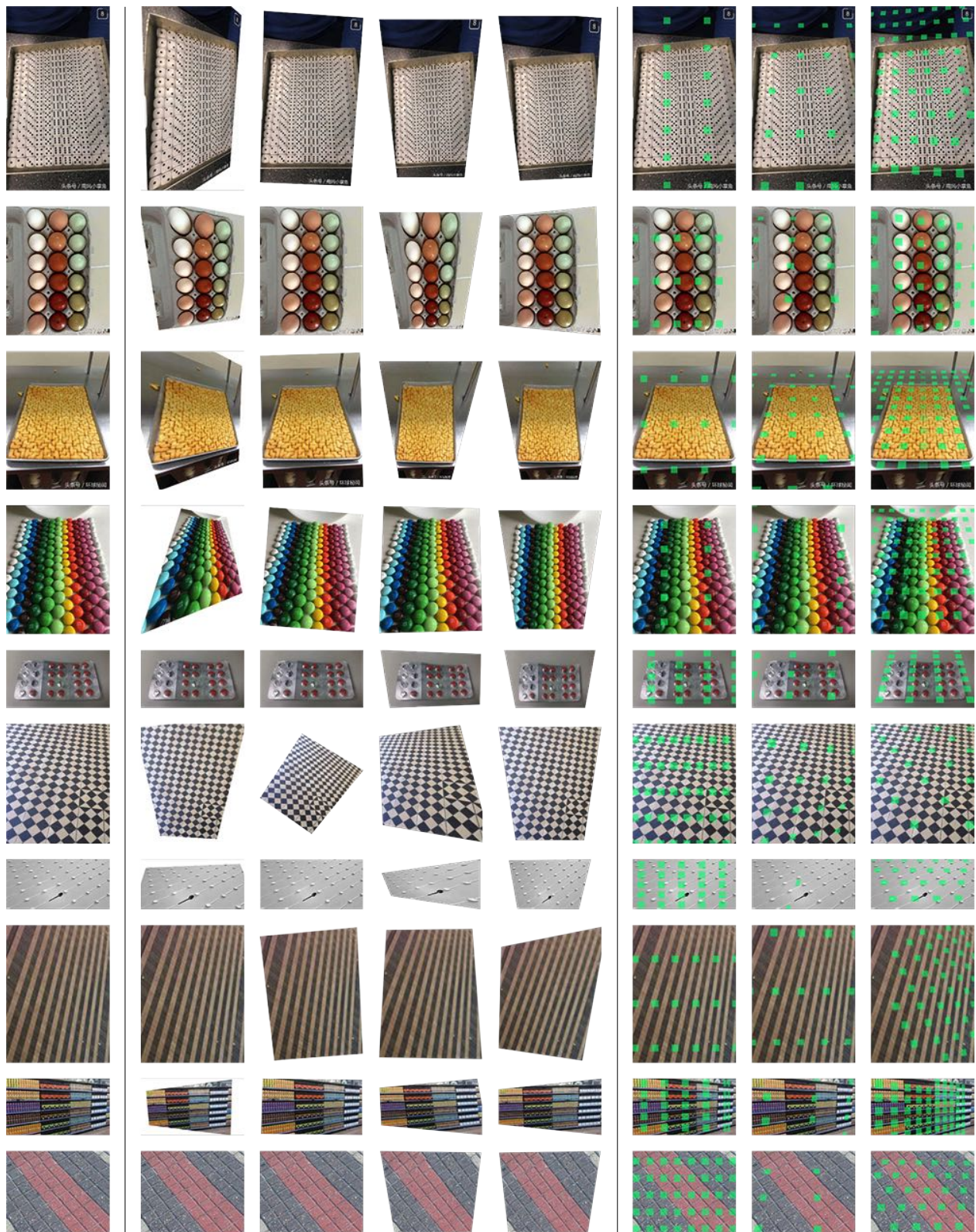

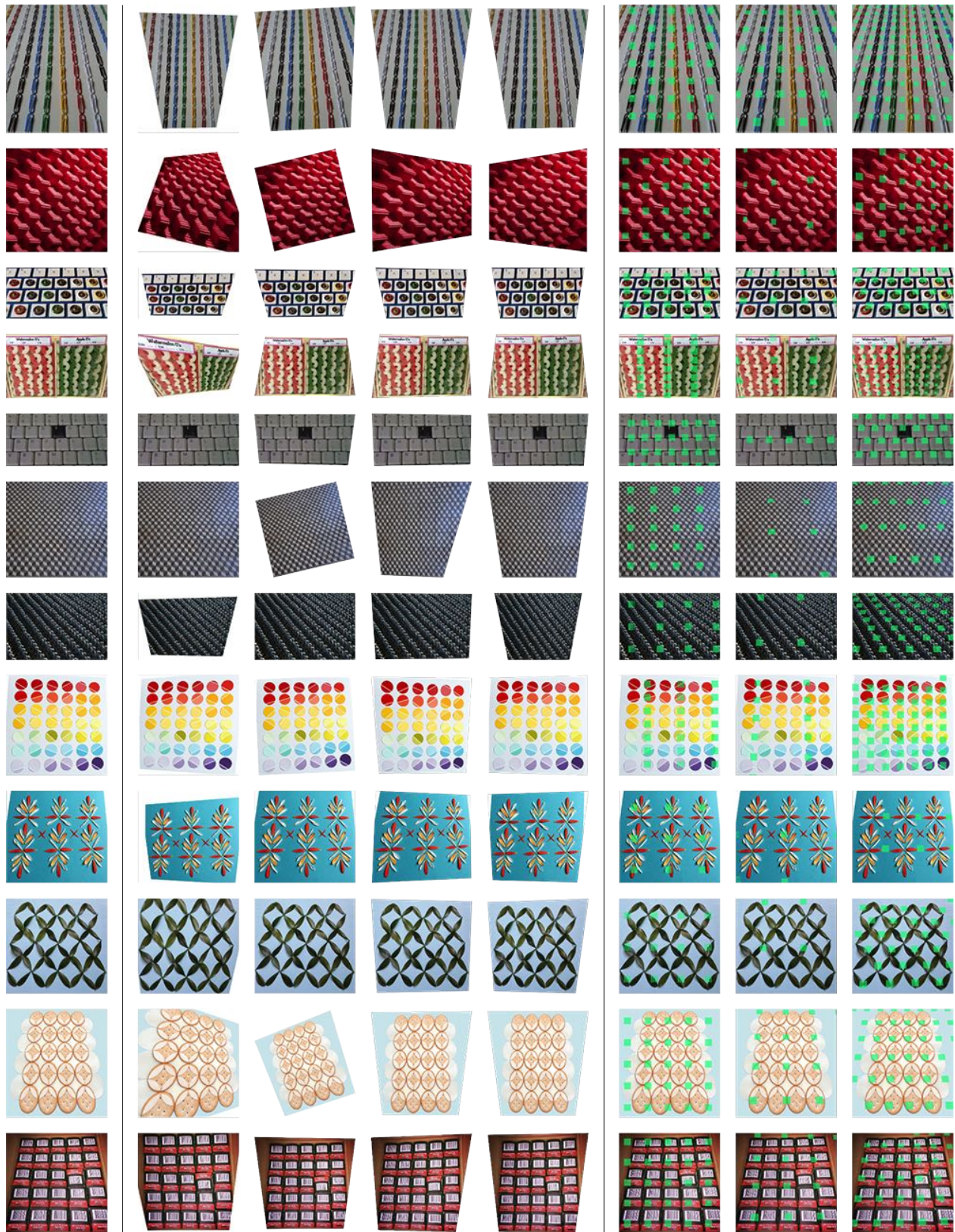

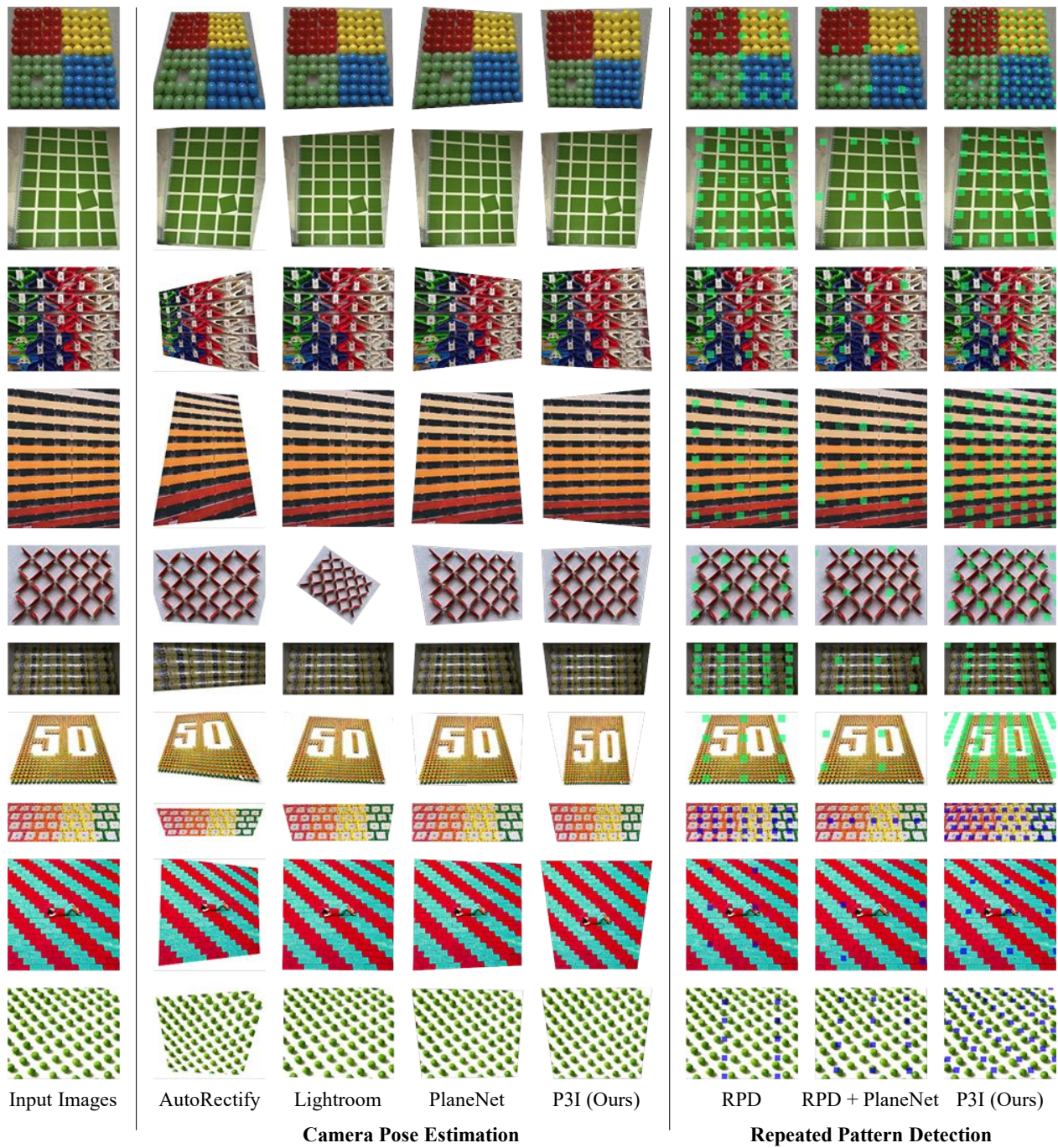

Figure 1: Qualitative results for camera pose estimation and repeated pattern detection. On the left, P3I outperforms both Lightroom and PlaneNet [5] on camera pose estimation. Results are visualized by performing an perspective correction based on the estimated parameters. On the right, we show that P3I can perform perspective-aware repeated pattern detection, which outperforms the baseline methods RPD [4] and RPD + PlaneNet.

**(a)** Image with objects placed in lattice patterns.

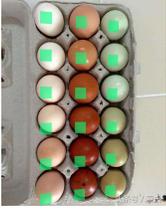
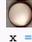
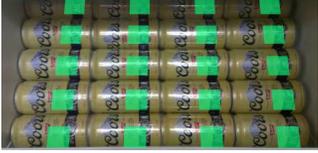
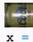
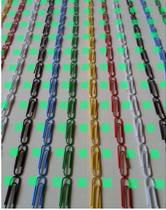
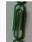
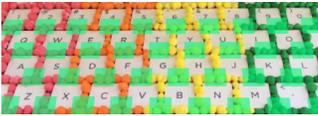
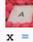
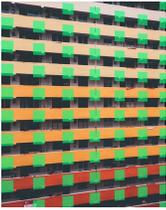
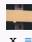
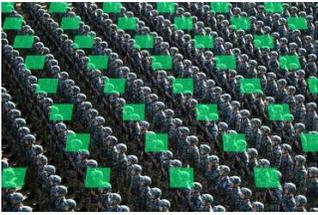
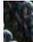
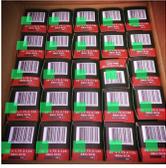
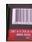
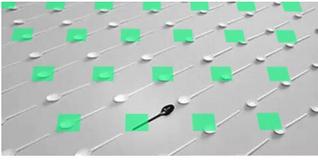
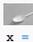
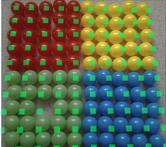
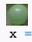
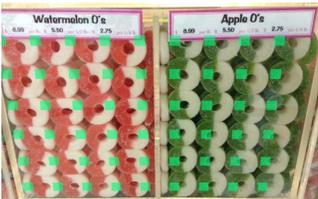
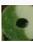

**(b)** Image with objects placed in circle patterns.

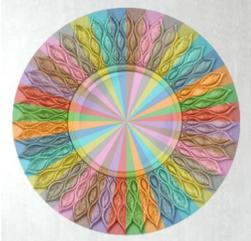
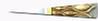
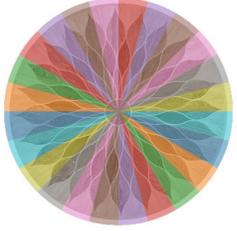
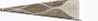

Figure 2: Examples of the induced perspective plane programs from our dataset NRPP. We overlay the centroids of the repeated objects as green squares on the original images for better visualization. We show the induced programs for both images with lattice patterns and images with circular patterns.

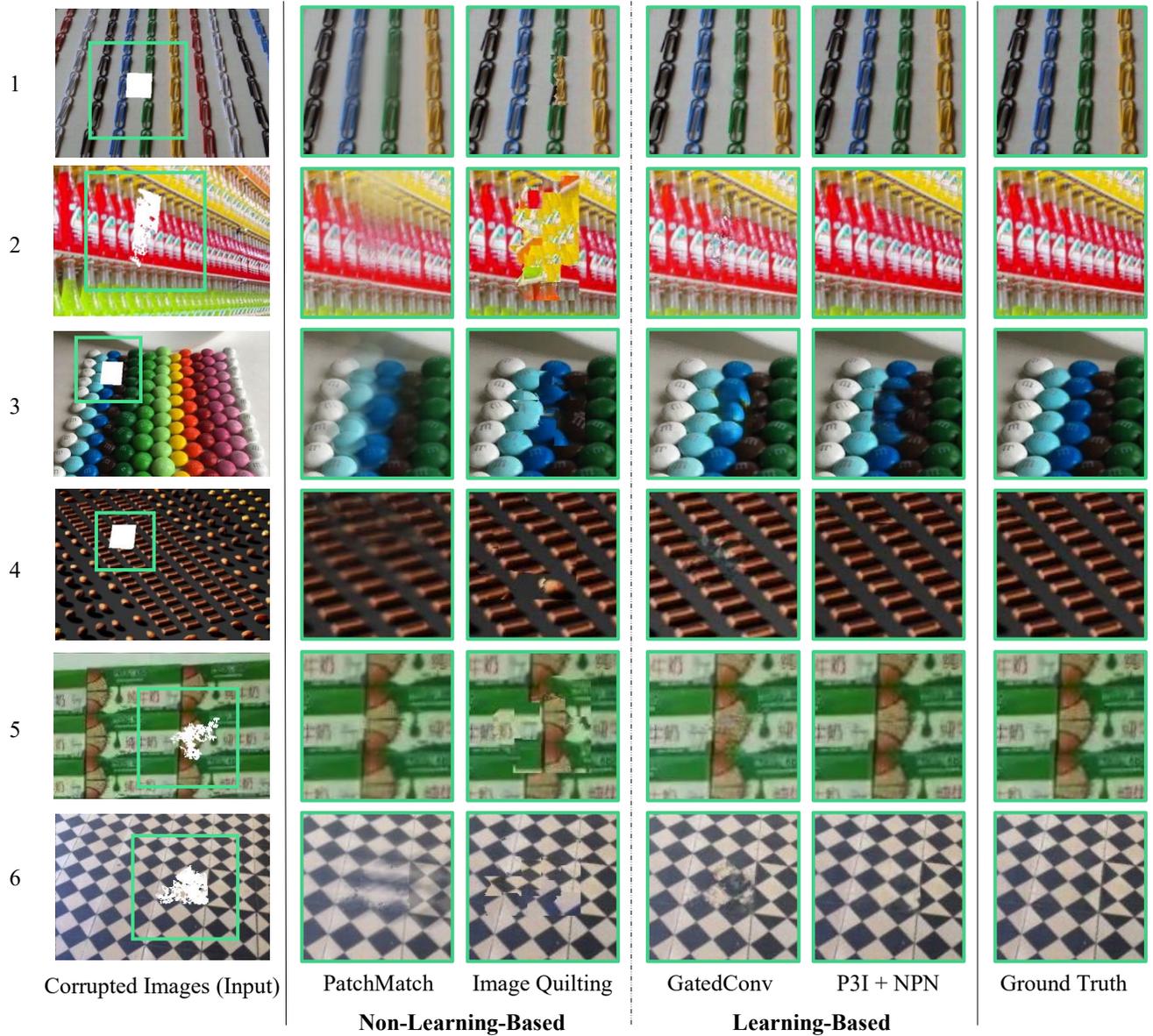

Figure 3: P3I-guided NPNs perform image inpainting in a perspective-aware fashion. Results produced by NPNs are sharp and consistent with the global structure. We compare our results with both non-learning-based methods (PatchMatch [2] and Image Quilting [3]) and a learning-based method (GatedConv [9]).



\end{document}